\documentclass[]{agibot}
\usepackage{enumitem}
\usepackage[toc,page,header]{appendix}

\usepackage[utf8]{inputenc} 
\usepackage[T1]{fontenc} 
\usepackage{hyperref} 
\usepackage{url}
\usepackage{array} 
\usepackage{booktabs}
\usepackage{amsfonts} 
\usepackage{nicefrac}
\usepackage{microtype} 
\usepackage{xcolor} 
\usepackage{xspace}
\usepackage{bm}
\usepackage{bbm}
\usepackage{bbding}
\usepackage{tabularx}
\usepackage{textcomp}
\usepackage{amssymb}
\usepackage{amsmath}
\usepackage{mathtools}
\usepackage{amsthm}
\usepackage{multirow}
\usepackage{makecell}
\usepackage{color}
\usepackage{colortbl}
\usepackage{adjustbox}
\usepackage{caption}
\usepackage{graphicx}
\usepackage{wrapfig}
\usepackage{multicol}
\usepackage{lipsum}
\usepackage{cleveref}

\crefname{figure}{Figure}{Figures}
\Crefname{figure}{Figure}{Figures}
\crefname{table}{Table}{Tables}
\Crefname{table}{Table}{Tables}

\definecolor{myyellow}{RGB}{255,192,0}
\definecolor{mygreen}{RGB}{107,170,64}
\definecolor{mywrite}{RGB}{255,227,132}

\newcommand{\method}{Ego-CEVR\xspace}

\newcolumntype{Y}{>{\centering\arraybackslash}X}
\newcolumntype{L}{>{\raggedright\arraybackslash}X}

\title{Ego-CEVR: Learning Cross-Embodiment Visual Representations from Human Egocentric Videos}

\author[1,2]{\textbf{Keru Zhou}}
\author[3]{\textbf{Zhirui Fang}}
\author[2]{\textbf{Xinrun Xu}}
\author[4]{\textbf{Yaopeng Lou}}
\author[5]{\textbf{Zeyuan He}}
\author[1]{\textbf{\\Zihao Zhu}}
\author[1]{\textbf{Xinze Niu}}
\author[1]{\textbf{Chen Wang}}
\author[1]{\textbf{Tianfu Li}}
\author[2]{\textbf{Samith Va}}
\author[6]{\textbf{Xiyue Zhang}}
\author[2]{\textbf{\\Bin Qian}}
\author[2]{\textbf{Junwei Li}}
\author[7]{\textbf{Wenjia Ding}}
\author[1]{\textbf{Yicheng Xu}}
\author[1]{\textbf{Bowen Jiang}}
\author[1]{\textbf{Wenhao Wang}}
\author[3]{\textbf{\\Ziwei Wang}}
\author[1,\dag]{\textbf{Chang Zhou}}
\author[1]{\textbf{Li Liu}}
\author[2,*]{\textbf{Xiu Li}}
\author[1,*]{\textbf{Maoqing Yao}}

\affiliation[1]{Maniformer}
\affiliation[2]{Tsinghua University}
\affiliation[3]{Nanyang Technological University}
\affiliation[4]{Huazhong University of Science and Technology}
\affiliation[5]{The Chinese University of Hong Kong, Shenzhen}
\affiliation[6]{Zhejiang University}
\affiliation[7]{Harbin Institute of Technology}

\firstfig[width=\linewidth][\textwidth]{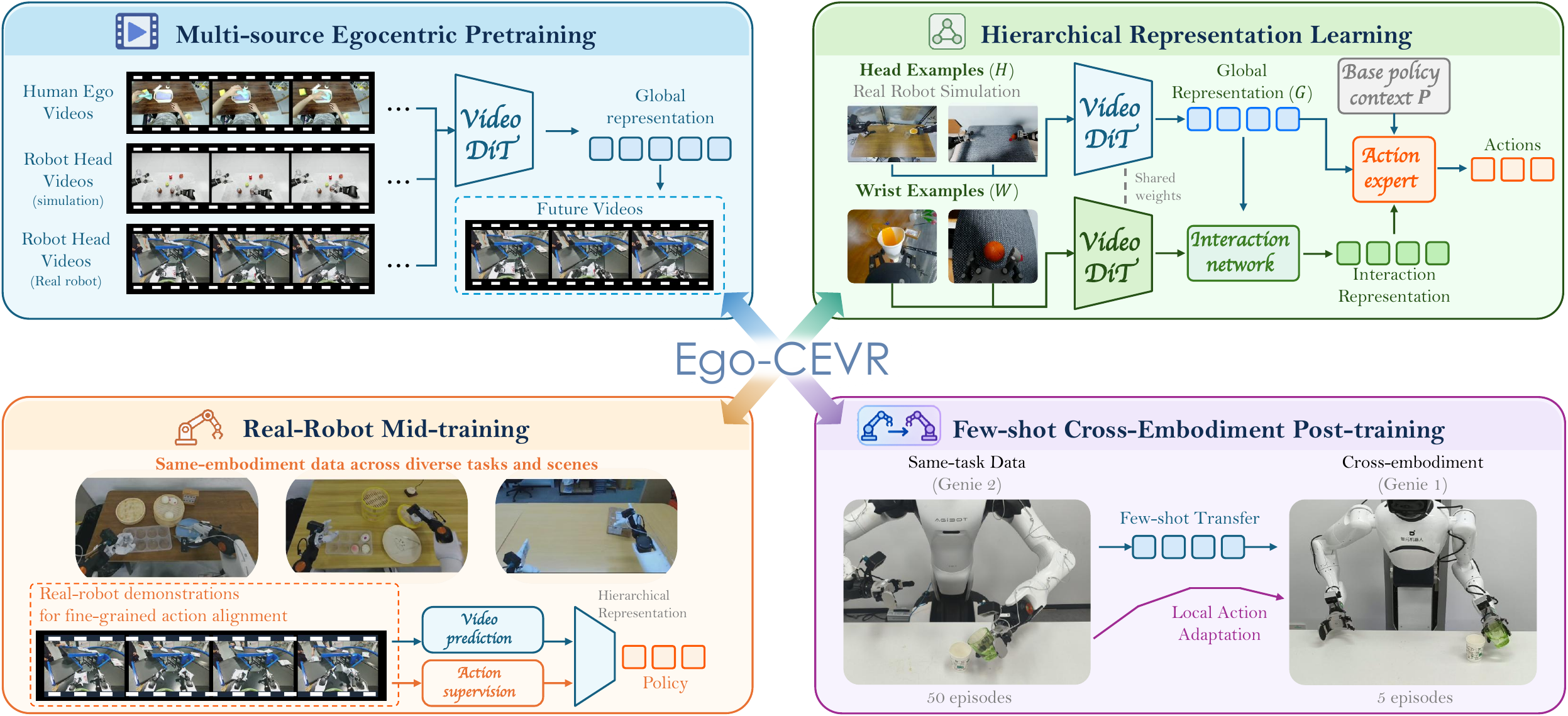}
{\textbf{Overview of Ego-CEVR.} Our framework combines global task context learned from multi-source egocentric videos with local interaction information grounded in wrist observations and robot actions. Training proceeds through video pretraining for global task context, mid-training across scenes for fine-grained action alignment, and post-training for policy adaptation to the target embodiment and scene. The illustrated few-shot setting uses demonstrations of the same task from a target robot for cross-embodiment adaptation.}
{fig:fig1}

\abstract{
Egocentric videos offer a low-cost, scalable source of manipulation experience for robot learning.
However, existing methods often fail to tailor their use of egocentric videos and robot demonstrations to the differences in their formats and information content.
We present \method, a data-native framework for hierarchical visual representation learning that assigns distinct learning objectives to global and interaction representations.
The global module learns task context through future video prediction, while the interaction module uses this context to interpret wrist observations and learns local execution information through robot action supervision. 
The distinct learning objectives enable independent pretraining on large-scale human and robot egocentric videos to learn task context, and training on real-robot demonstrations to align the learned representations with fine-grained robot actions.
Experiments show leading performance on LIBERO-Plus and RoboTwin 2.0, while real-robot evaluations demonstrate generalization across tasks and embodiments.
}

\definecolor{BlockC}{gray}{0.98}  
\definecolor{BlockA}{RGB}{191,211,230}
\definecolor{BlockB}{RGB}{199,233,192}

\begingroup

\footnotetext[1]{Corresponding authors: Xiu Li and Maoqing Yao.}
\footnotetext[2]{Project leader: Chang Zhou.}
\endgroup

\begin{document}
\maketitle

\section{Introduction}
\label{sec:intro}

As egocentric video becomes increasingly available for embodied learning~\cite{kareer2025egomimic}, an important question is how to turn this low-cost, scalable source of data into robot manipulation capabilities. Robot learning also draws on Universal Manipulation Interface (UMI) data~\cite{chi2024universal} and demonstrations from different robot embodiments~\cite{o2024open}. These data sources differ in their observations and forms of supervision, and can therefore play different roles in learning~\cite{ye2026data}. Making effective use of them requires data-native models whose learning objectives match the information and supervision each source provides~\cite{ye2026data,sun2026ego4wam,liu2026roboicl,cheng2026context}.

Vision-language-action (VLA) policies and predictive world models offer complementary capabilities for learning from the different forms of supervision available across these data sources. 
VLA policies transfer visual and linguistic priors to robot control~\cite{zitkovich2023rt,kim2024openvla,intelligence2025pi_}, yet semantic generalization does not always translate into faithful execution on unseen tasks~\cite{chen2026instructions,fang2026vision,zhang2026flatness}.
Predictive world models learn how scenes evolve during manipulation~\cite{hu2024video,liao2026genie}, but using this knowledge for precise local control remains challenging~\cite{liu2026ge}.
Therefore, further research is needed to effectively integrate these capabilities~\cite{miao2026jepa,chen2026uniwam}.
To benefit from heterogeneous video and robot data, such integration must clarify which representations can be learned without action annotations and how robot demonstrations can be used to align these representations with robot actions.

Reusing manipulation experience from humans or other robots requires handling differences in morphology, action space, and local interaction geometry~\cite{kareer2025emergence,ye2026world}.
Despite differences in embodiment, similar manipulation tasks often involve similar changes in the spatial and functional relationships between objects.
Human egocentric videos and robot head-camera videos can therefore provide supervision for learning task-level visual representations~\cite{kareer2025egomimic,sun2026ego4wam}.
In contrast, wrist observations provide a closer view of the end effector and its target, revealing local cues for action alignment and contact. 
Guided by these task-level representations, local interaction features should combine task-level action semantics with spatial information from wrist observations.
Robot demonstrations provide supervision to align these features with actions.

Following this principle, we introduce \method, a data-native framework for hierarchical visual representation learning (\cref{fig:fig1}).
It learns global task representations from human egocentric and robot head-camera videos through future video prediction, and local interaction representations from wrist observations under robot action supervision. 
Wrist features query the global representation to incorporate task context into local interaction features.
The global representation can be learned without wrist observations or robot action supervision, allowing the global module to be pretrained independently on large-scale video data.

Experiments on LIBERO~\cite{liu2023libero}, LIBERO-Plus~\cite{fei2026libero} and  RoboTwin 2.0~\cite{chen2025robotwin} show competitive manipulation performance and improved robustness. Real-robot experiments further show that human egocentric videos improve task performance and support task transfer.
The results also suggest that the learned hierarchical representations support few-shot cross-embodiment knowledge transfer. Ablations examine the contributions of the hierarchical structure, view-specific learning objectives, and different video sources.

Our main contributions are:
\begin{itemize}
    \item We propose a data-native framework that organizes learning from multi-source egocentric videos and robot demonstrations into three stages, with pretraining for global task context, mid-training for fine-grained action alignment, and post-training for policy adaptation.

    \item We design a hierarchical architecture that learns complementary global and interaction representations. Global representations provide task-level action semantics and guide interaction representations that incorporate local spatial information from wrist observations.

    \item We enable few-shot cross-embodiment adaptation by combining task knowledge learned from video with a few target-robot demonstrations for action alignment.
\end{itemize}

\section{Related Work}
\label{sec:related_work}

\begin{figure*}[t]
    \centering
    \includegraphics[width=\textwidth]{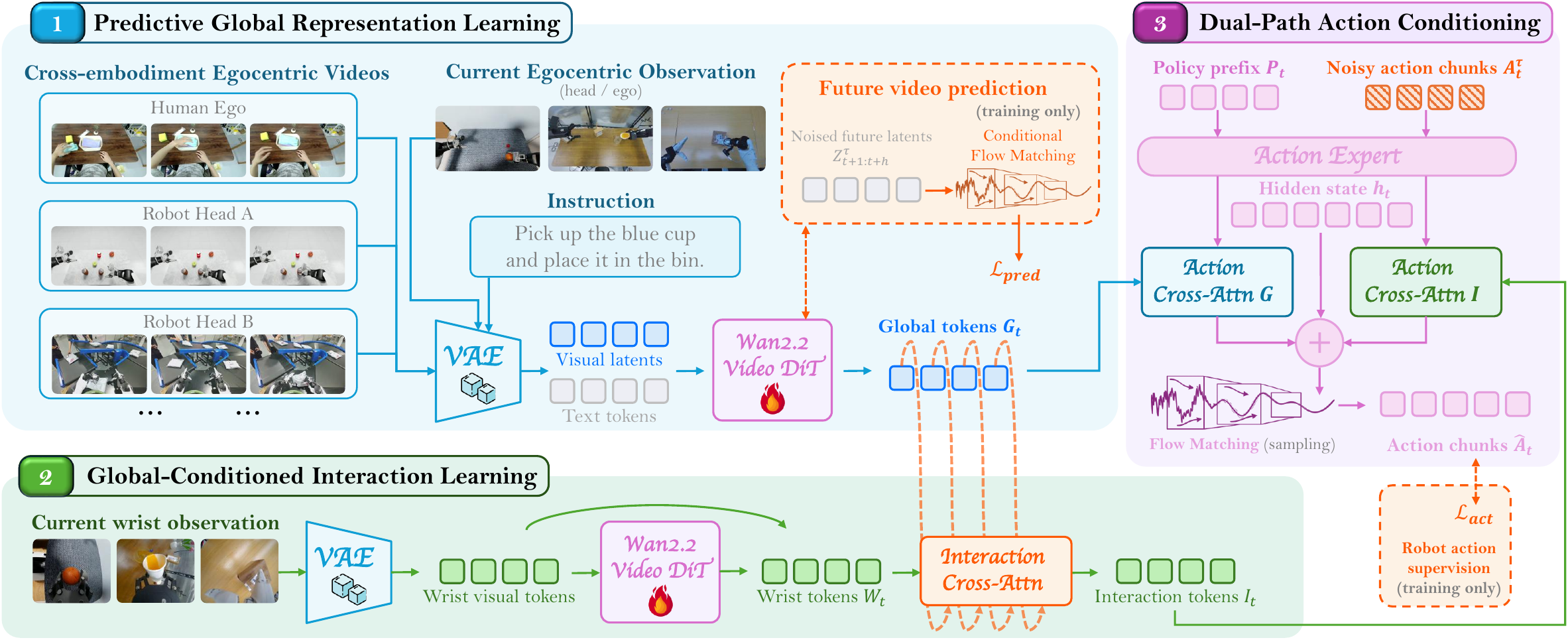}
    
    \caption{\textbf{Architecture of Ego-CEVR.} The global module learns task-level representations from egocentric videos through future video prediction. Guided by the global representation, the interaction module produces local interaction representations from wrist observations. The action expert uses both representations together with the base-policy context for action generation.}
    \label{fig:fig2}
\end{figure*}

\paragraph{Vision-Language-Action models.}
VLA models transfer pretrained visual and linguistic knowledge to robot control, achieving substantial progress in semantic generalization, multitask manipulation, and deployment across environments~ \cite{zitkovich2023rt,kim2024openvla,black2024pi_0,intelligence2025pi_,bjorck2025gr00t,team2025gemini}. However, on unseen tasks, VLA models may repeat behaviors learned during training and fail to follow the current instruction~\cite{ye2026world}. This motivates incorporating future visual prediction into policy learning, providing supervision on how scenes change during manipulation and enabling learning from video experience beyond action-labeled robot demonstrations~\cite{wu2024unleashing,cheang2024gr}.

\paragraph{World action models.}
Video foundation models provide rich spatiotemporal priors for learning manipulation processes and their visual consequences~\cite{wan2025wan,agarwal2025cosmos}. Building on these priors, world action models (WAMs) guide action generation using predicted visual states or video-model representations~\cite{hu2024video,liao2026genie,li2026causal,ye2026world}. However, visually plausible predictions may fail to preserve the relative positions, alignment, and contact relationships required for precise execution~\cite{liu2026ge}. On unseen tasks and scenes, prediction errors may further propagate into action errors~\cite{guo2026video}. Recent work therefore focuses on learning predictive representations that retain information relevant to action control~\cite{zhang2026native}. Fast-WAM~\cite{yuan2026fast} shows that video prediction during training can improve control representations without requiring explicit future-video generation at deployment. VPP2~\cite{guo2026video} improves generalization in video prediction and action generation through manipulation-video pretraining and subsequent adaptation for robot control. These studies highlight the importance of combining manipulation knowledge learned from video with the local spatial information needed for precise execution.

\paragraph{Integrating VLA models and world models.}
Recent work combines vision-language understanding with predictive world modeling for robot control~\cite{bi2026motus,chen2026uniwam,zheng2025flare,miao2026jepa,sun2026vla}. In terms of architecture, Motus~\cite{bi2026motus} and UniWAM~\cite{chen2026uniwam} integrate semantic understanding, visual generation, and action prediction within unified architectures. In terms of representations, FLARE~\cite{zheng2025flare} aligns policy features with latent representations of future observations, while VLA-JEPA~\cite{sun2026vla} learns representations through future latent-state prediction before action learning. JEPA-VLA~\cite{miao2026jepa} incorporates pretrained video-predictive embeddings into VLA models through gated cross-attention. 
However, less attention has been paid to organizing representations according to the information provided by different camera views. VPP2~\cite{guo2026video} jointly predicts multiple views and uses video-model representations for action generation, without explicitly assigning distinct learning objectives to different views. This leaves open how video prediction and robot action supervision should support task-level and local interaction learning from head and wrist observations, and how these representations should be connected for precise execution.

\paragraph{Learning from human egocentric videos.}
Human egocentric videos capture diverse object interactions and task variations, providing coverage that is difficult to obtain through robot teleoperation alone~\cite{sun2026ego4wam}. Early work uses such videos to learn transferable visual representations~\cite{nair2022r3m} and rewards~\cite{ma2022vip}. Subsequent methods transfer human manipulation experience to robot policies through human--robot data alignment, hand-pose supervision, or action retargeting~\cite{kareer2025egomimic,yang2025egovla,zheng2026egoscale}. These approaches demonstrate the value of human experience, but obtaining motion annotations or aligned human--robot data introduces additional data collection and processing requirements~\cite{sun2026ego4wam}. 
EgoWAM~\cite{li2026egowam} examines how different visual prediction targets affect the transfer of human experience to robot control. Ego4WAM~\cite{sun2026ego4wam} studies the effects of human--robot data alignment and supervision, showing that video-only supervision can provide an effective foundation for subsequent video-action training. Trained on semantically matched pairs of human videos and robot trajectories, Zero-WAM~\cite{zhou2026zero} uses human videos as in-context task specifications to execute unseen tasks without parameter updates. Together, these studies highlight the importance of connecting transferable task knowledge from human videos with local execution knowledge learned through robot action supervision.

\section{Method}
\label{sec:method}

\subsection{Framework Overview}

We use egocentric data to refer to human egocentric videos and robot head-camera videos.
For video pretraining, only egocentric videos with corresponding language instructions are used, without paired wrist observations or robot action annotations.
Robot demonstrations, in contrast, include language instructions, observations from a head camera and two wrist cameras, robot states and action annotations.
We denote the egocentric observation at time $t$ by $I_t^H$, the pair of wrist observations by $I_t^W$, the action sequence by $A_t$, and the language instruction by $\ell$.

As illustrated in \cref{fig:fig2}, \method consists of a global video representation module, an interaction representation module, and a representation-augmented action expert. The global module learns from egocentric video through future prediction and produces a representation $G_t$. The interaction module uses wrist observations and $G_t$ to produce an interaction representation $I_t$. The action expert receives both representations through separate attention branches and uses them alongside the base-policy context for action generation.

Training proceeds in three stages.
We first pretrain the global module on multi-source egocentric videos to learn task context through future video prediction.
Mid-training then uses real-robot demonstrations collected across different scenes to learn fine-grained action alignment.
Finally, post-training fine-tunes the policy using demonstrations from the target embodiment and scene, supporting few-shot cross-embodiment adaptation.
The following subsections describe the representation learning process and how the learned representations are used for action generation.


\subsection{Predictive Global Representation Learning}

We use the diffusion transformer (DiT)~\cite{peebles2023scalable} of the pretrained Wan2.2 video model~\cite{wan2025wan} as our visual backbone. Given the current egocentric observation $I_t^H$ and language instruction $\ell$, the encoder $E$ of a frozen variational autoencoder (VAE)~\cite{kingma2013auto} maps the observation to a latent $z_t = E(I_t^H)$. The language-conditioned video DiT processes $z_t$ at a zero noise timestep. We extract attention value tokens from selected intermediate layers and concatenate them along the token dimension to form the global representation $G_t$. The resulting tokens retain spatial information from these layers for interaction modeling and action prediction. At inference, extracting $G_t$ does not require generating future video.

We supervise the video backbone through conditional flow matching~\cite{lipman2022flow} over future video latents. The frozen VAE encoder $E$ maps the current egocentric frame to $z_t$ and the subsequent video frames to future latents $Z_{t+1:t+h}$, where $h$ denotes the prediction horizon in latent timesteps. During training, we keep $z_t$ clean and construct noisy future latents through linear interpolation as
\begin{equation}
    \begin{aligned}
        Z_{t+1:t+h}^{\tau} &= (1-\tau)Z_{t+1:t+h}+\tau\epsilon, \\
        \epsilon &\sim \mathcal{N}(0,I), \quad \tau \in [0,1].
    \end{aligned}
\end{equation}
Given the noisy future latents $Z_{t+1:t+h}^{\tau}$ and noise timestep $\tau$, the video DiT predicts the vector field $u_v$, conditioned on $z_t$ and the language instruction $\ell$. The prediction objective is
\begin{equation}
    \mathcal{L}_{\mathrm{pred}}
    = \mathbb{E}_{\tau,\epsilon}
    \bigl(
        w(\tau) \left\|u_v-(\epsilon-Z_{t+1:t+h})\right\|^2
    \bigr),
\end{equation}
where $w(\tau)$ is the timestep-dependent loss weight and $\|\cdot\|$ is the Euclidean norm. By modeling visual changes during manipulation, $\mathcal{L}_{\mathrm{pred}}$ trains the video backbone to produce global representations from current egocentric observations.

During pretraining, we update only the video DiT and keep the VAE frozen.
We then use the resulting DiT weights to initialize the video backbone for mid-training.

\subsection{Global-Conditioned Interaction and Action}

Given the current wrist observations $I_t^W$, we obtain wrist features $W_t$ using the frozen VAE encoder and shared video DiT introduced in the previous section. The wrist observations are processed in a separate forward pass using the same cross-layer feature readout. The interaction network $F_I$ then attends to the global representation $G_t$ to incorporate scene and task context into the wrist features, yielding an interaction representation defined as
\begin{equation}
    I_t = F_I(W_t,G_t).
\end{equation}

The resulting interaction representation allows local wrist cues for approach, alignment, and contact to be interpreted in the context of the current task and scene.

For action generation, the vision-language backbone encodes tokens from the current head and wrist observations, the language instruction, and robot state into the prefix context $P_t$~\cite{intelligence2025pi_}.
Conditioned on $P_t$ and the flow time $\tau$, the action expert processes the noisy action sequence and produces final-layer hidden states $h_t$.
We use two independent cross-attention branches, $\mathcal{C}_G$ and $\mathcal{C}_I$, to attend to $G_t$ and $I_t$, respectively.
Their outputs are weighted by $\alpha$ and $\beta$ and added to $h_t$ to obtain
\begin{equation}
    \widetilde h_t = h_t + \alpha \cdot \mathcal{C}_G(h_t, G_t)
                    + \beta \cdot \mathcal{C}_I(h_t, I_t).
\end{equation}
Both branches use $h_t$ as queries. Their keys and values come from $G_t$ and $I_t$, respectively. The learnable scaling parameters $\alpha$ and $\beta$ are initialized to zero. The fused states $\widetilde h_t$ are mapped through the output projection to the action vector field $u_a$. At inference, action sequences are generated by integrating this vector field from Gaussian noise.

For flow-matching action training~\cite{intelligence2025pi_}, we construct the noisy action sequence from the ground-truth sequence $A_t$ as
\begin{equation}
        A_t^\tau = (1-\tau)A_t+\tau\epsilon_a, 
        \quad \epsilon_a \sim \mathcal{N}(0,I), \tau \in [0,1].
\end{equation}
The predicted action vector field is
\begin{equation}
    \hat{u}_t^\tau
    = u_a(A_t^\tau,\tau \mid P_t,G_t,I_t).
\end{equation}

We train the action policy with a squared $L_2$ loss given by
\begin{equation}
    \mathcal{L}_{\mathrm{act}}
    = \mathbb{E}_{\tau,\epsilon_a}
    \left\|
        \hat{u}_t^\tau-(\epsilon_a-A_t)
    \right\|^2.
\end{equation}
This loss trains the action expert and the two attention branches to use the global and interaction representations for action prediction. It also provides action supervision for the interaction network $F_I$, encouraging $I_t$ to retain information relevant to local execution.

During training on robot demonstrations, we combine action supervision with egocentric video prediction using the overall objective given by
\begin{equation}
    \mathcal{L}_{\mathrm{total}}
    =
    \mathcal{L}_{\mathrm{act}}
    +
    \lambda_{\mathrm{pred}}\mathcal{L}_{\mathrm{pred}},
\end{equation}
where $\lambda_{\mathrm{pred}}$ controls the relative contribution of the video prediction loss. The action-conditioning representations $P_t$, $G_t$, and $I_t$ are constructed from current observations and task conditions, while future egocentric frames are used only to compute $\mathcal{L}_{\mathrm{pred}}$.

\section{Experiments}
\label{sec:experiments}

\begin{wraptable}{l}{0.48\textwidth}
\centering

\caption{Manipulation performance on the four LIBERO task suites. Results are success rates (\%). Bold indicates the best result in each column.}
\label{tab:libero}
\small
\setlength{\tabcolsep}{3.2pt}
\resizebox{\linewidth}{!}{%
\begin{tabular}{lccccc}
\toprule
Method & Spatial & Object & Goal & Long & Avg. \\
\midrule
$\pi_{0}$~\cite{black2024pi_0} & 96.8 & 98.8 & 95.8 & 85.2 & 94.2 \\
$\pi_{0.5}$~\cite{intelligence2025pi_} & 98.8 & 98.2 & 98.0 & 92.4 & 96.9 \\
OpenVLA-OFT~\cite{kim2025fine} &  97.6 & 98.4 & 97.9 & 94.5 & 97.1 \\
VLA-JEPA~\cite{sun2026vla} & 96.2 & 99.6 & 97.2 & \textbf{95.8} & 97.2 \\
Fast-WAM~\cite{yuan2026fast} & 98.2 & \textbf{100} & 97.0 & 95.2 & 97.6 \\
\midrule
Ego-CEVR (ours) & \textbf{99.2} & \textbf{100} & \textbf{98.2} & 94.8 & \textbf{98.1} \\
\bottomrule
\end{tabular}}
\end{wraptable}

We evaluate \method in simulation and on real robots. We first compare manipulation performance and robustness with representative baselines. Controlled ablations then examine the hierarchical structure, global-to-interaction guidance, and view-specific learning objectives. Finally, real-robot experiments assess learning from different video sources, video-guided task transfer, and few-shot cross-embodiment adaptation.

\subsection{Experimental Setup}
\label{sec:eval_setup}

\paragraph{Simulation setup.}
We evaluate manipulation performance on the Spatial, Object, Goal, and Long suites of LIBERO~\cite{liu2023libero}. On LIBERO-Plus~\cite{fei2026libero}, we use the policy trained on LIBERO to evaluate robustness to seven types of perturbations, including Camera, Robot, Language, Light, Background, Noise, and Layout. On RoboTwin 2.0~\cite{chen2025robotwin}, we train on clean data and evaluate bimanual manipulation performance and robustness to environmental variations under clean-to-clean (Easy) and clean-to-random (Hard) settings. All benchmark evaluations follow their standard protocols, with task success rate (\%) as the evaluation metric.

\begin{table*}[t]
\centering
\caption{Robustness evaluation on LIBERO-Plus across seven perturbation types. Results are success rates (\%). Bold indicates the best result in each column.}
\label{tab:libero_plus}
\small
\setlength{\tabcolsep}{4.3pt}
\resizebox{\textwidth}{!}{%
\begin{tabular}{lcccccccc}
\toprule
Method & Camera & Robot & Language & Light & Background & Noise & Layout & Average \\
\midrule
$\pi_{0.5}$~\cite{intelligence2025pi_} & 75.8 & 79.4 & 83.3 & 95.5 & 95.0 & 89.6 & 87.0 & 85.7 \\
ACoT-VLA~\cite{zhong2026acot} & 72.6 & 82.6 & 87.5 & 97.7 & 96.5 & 87.8 & 86.6 & 86.6 \\
VLA-JEPA~\cite{sun2026vla} & 63.3 & 67.1 & 85.4 & 95.6 & 93.6 & 66.3 & 85.1 & 79.5 \\
JEPA-WAM~\cite{lin2026jepa} & 79.2 & 59.2 & 68.2 & 93.3 & 94.6 & 83.6 & 76.1 & 79.2 \\
Abot-M0.5~\cite{chen2026abot} & 70.5 & \textbf{87.4} & \textbf{88.6} & 94.0 & 89.7 & 75.5 & 85.2 & 83.4 \\
Being-H0.7~\cite{luo2026being} & 82.0 & 59.0 & 82.8 & \textbf{97.8} & 90.0 & 93.5 & \textbf{88.5} & 84.8 \\
LingBot-VA~\cite{li2026causal} & 40.9 & 83.0 & 86.4 & 82.3 & 53.1 & 64.4 & 76.2 & 69.5 \\
Cosmos-Policy~\cite{kim2026cosmos} & 75.8 & 63.3 & 81.7 & 96.5 & 88.9 & 92.7 & 82.2 & 82.2 \\
Fast-WAM~\cite{yuan2026fast} & 17.7 & 45.3 & 60.1 & 52.2 & 22.8 & 28.5 & 61.2 & 41.3 \\
\midrule
Ego-CEVR (ours) & \textbf{91.1} & 78.3 & 86.6 & 97.7 & \textbf{98.8} & \textbf{95.4} & 85.7 & \textbf{90.2} \\
\bottomrule
\end{tabular}}
\end{table*}

\paragraph{Real-robot setup.}
We conduct real-robot experiments on the AgiBot G1 and G2 platforms. For video-source comparisons, we keep robot demonstrations fixed and compare demonstrations-only training with settings that add 10 h of robot head-camera or human egocentric videos. Task transfer is evaluated from Water Pouring to Block Stacking on G2. Few-shot cross-embodiment adaptation is evaluated on Water Pouring from G2 to G1, using 50 G2 demonstrations and five G1 demonstrations.

\paragraph{Comparisons and controls.}
We compare \method with representative VLA policies and world-model-based manipulation methods. Simulation baseline results are taken from the original papers or other published studies. For real-robot comparisons, we train the baselines using the same robot demonstrations and evaluate them under the same conditions as \method. We also conduct controlled ablations within our framework to examine the contributions of the architecture and learning objectives. Across these variants, we keep video sources, robot demonstrations, camera inputs, base-policy initialization, and evaluation settings fixed. The ablation variants are described in the corresponding subsection. Detailed data configurations, training hyperparameters, and evaluation procedures are provided in the appendix.

\subsection{Manipulation and Robustness}
\label{sec:manipulation_robustness}

\paragraph{LIBERO.}
As shown in \cref{tab:libero}, Ego-CEVR achieves the highest average success rate among the listed methods at 98.1\%, exceeding Fast-WAM~\cite{yuan2026fast} and $\pi_{0.5}$~\cite{intelligence2025pi_} by 0.5 and 1.2 percentage points, respectively. It achieves the highest success rates on Spatial and Goal at 99.2\% and 98.2\%, respectively, and ties with Fast-WAM~\cite{yuan2026fast} on Object at 100.0\%. It also outperforms $\pi_{0.5}$~\cite{intelligence2025pi_} across all four suites, demonstrating competitive in-distribution manipulation performance.

\begin{wraptable}{l}{0.4\textwidth}
\centering
\caption{Bimanual manipulation on RoboTwin 2.0 under Easy and Hard settings. \method~Aug uses augmented head-view data to pretrain the global module. Results are success rates (\%). Bold indicates the best result in each column.}
\label{tab:robotwin}
\small
\setlength{\tabcolsep}{4pt}
\resizebox{\linewidth}{!}{%
\begin{tabular}{lccc}
\toprule
Method & Easy & Hard & Average \\
\midrule
$\pi_{0.5}$~\cite{intelligence2025pi_} & 70.7 & 46.0 & 58.4 \\
Fast-WAM~\cite{yuan2026fast} & 77.8 & 1.9 & 39.9 \\
UniWAM~\cite{chen2026uniwam} & 75.2 & 67.6 & 71.4 \\
WorldScape Policy 2.0~\cite{su2026worldscape} & \textbf{84.9} & 38.5 & 61.7 \\
4D-WAM~\cite{yang20264d} & 81.5 & 41.8 & 61.6 \\
JEPA-WAM~\cite{lin2026jepa} & 79.9 & 36.9 & 58.4 \\
X-WAM~\cite{guo2026unified} & 70.0 & 25.8 & 47.9 \\
Abot-M0~\cite{yang2026abot} & 57.4 & 30.4 & 43.9 \\
StarVLA~\cite{community2026starvla} & 46.5 & 3.2 & 24.8 \\
GalaxeaVLA~\cite{jiang2026galaxy} & 62.7 & 12.7 & 37.7 \\
\midrule
Ego-CEVR (ours)  & 73.7 & 70.7 & 72.2 \\
Ego-CEVR Aug (ours) & 78.2 & \textbf{74.3} & \textbf{76.2} \\
\bottomrule
\end{tabular}}
\end{wraptable}

\paragraph{LIBERO-Plus.}
As shown in \cref{tab:libero_plus}, \method achieves an overall success rate of 90.2\%, the highest among the listed methods and 3.6 percentage points above ACoT-VLA~\cite{zhong2026acot}. It achieves the highest success rates under camera, background, and noise perturbations at 91.1\%, 98.8\%, and 95.4\%, respectively. Under camera viewpoint perturbations, it outperforms $\pi_{0.5}$~\cite{intelligence2025pi_} by 15.3 percentage points. Together with the LIBERO results, these findings show that \method maintains competitive in-distribution manipulation performance while achieving stronger robustness to multiple types of perturbations.

\paragraph{RoboTwin 2.0.}
As shown in \cref{tab:robotwin}, Ego-CEVR achieves success rates of 73.7\% and 70.7\% under the Easy and Hard settings, respectively, yielding an average of 72.2\%. It outperforms all listed baselines in both average success rate and the Hard setting. The gains over UniWAM~\cite{chen2026uniwam} are 0.8 and 3.1 percentage points, respectively. The corresponding gains over $\pi_{0.5}$~\cite{intelligence2025pi_} are 13.8 and 24.7 percentage points. The success rate decreases by only 3.0 percentage points from Easy to Hard.
For Ego-CEVR Aug, we generate augmented head-view samples equal in number to those in the clean training set and use them to pretrain the global module before policy training. This variant achieves success rates of 78.2\% and 74.3\% under the Easy and Hard settings, respectively, with an average of 76.2\%, improving over Ego-CEVR by 4.0 percentage points on average. These results demonstrate robust bimanual manipulation under environmental distribution shifts and further performance gains from augmented video pretraining.

Overall, Ego-CEVR achieves competitive manipulation performance on LIBERO and demonstrates robustness to perturbations on LIBERO-Plus and out-of-distribution generalization on RoboTwin 2.0. We next conduct controlled ablations to examine the contributions of the hierarchical structure, global-to-interaction guidance, and view-specific learning objectives.

\begin{figure}[t]
    \centering
    \includegraphics[width=\columnwidth]{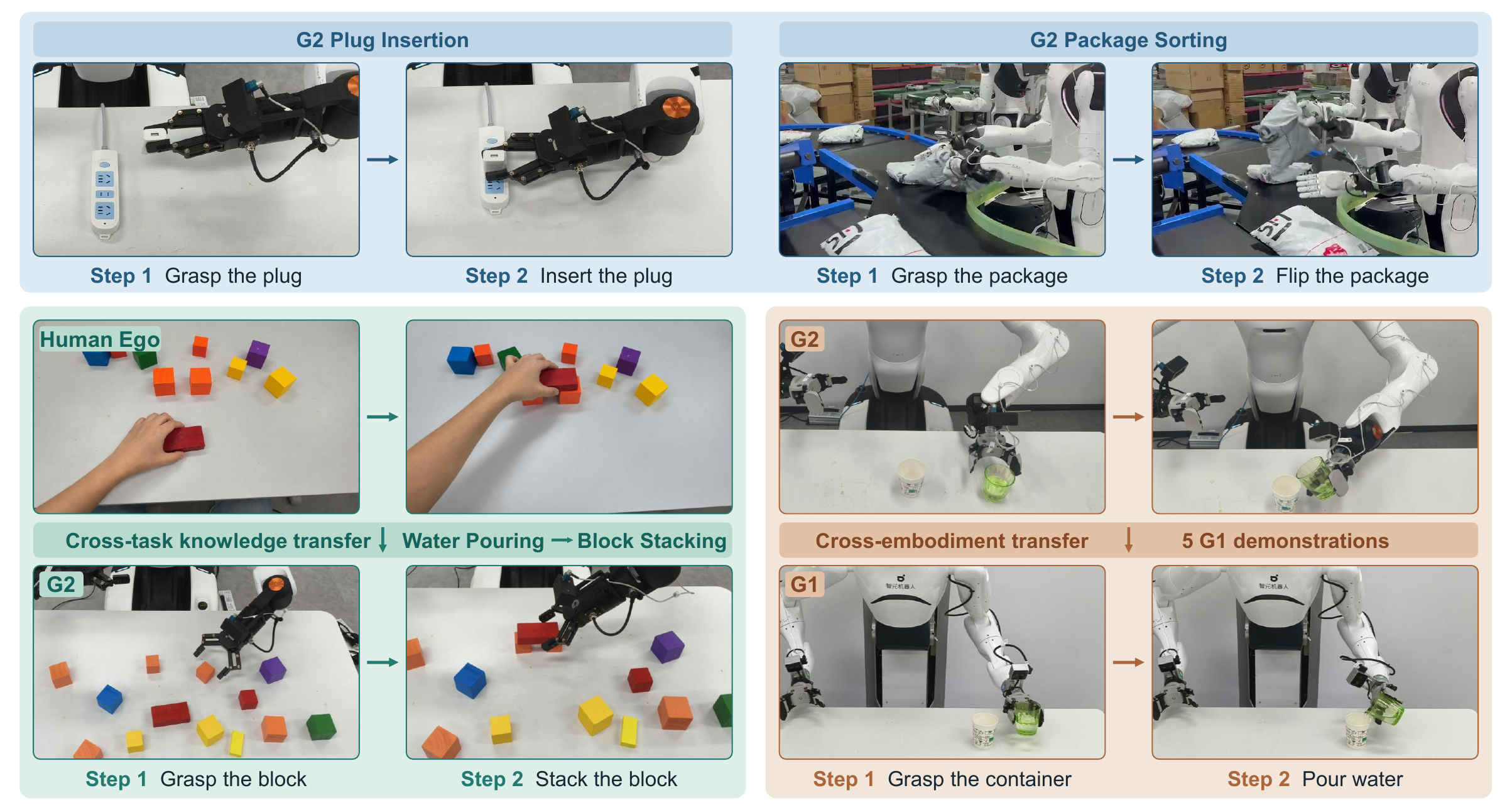}
    \caption{\textbf{Real-robot tasks and knowledge transfer.} Plug insertion and package sorting demonstrate manipulation on G2. For cross-task knowledge transfer from Water Pouring to Block Stacking on G2, human egocentric videos provide knowledge of the target task. On Water Pouring, 5 G1 demonstrations support cross-embodiment knowledge transfer from G2 to G1.}
    \label{fig:real}
\end{figure}

\subsection{Ablation Study on Hierarchical Representation}

\begin{table*}[ht]
\centering
\caption{Ablations of the hierarchical representation structure, global-to-interaction guidance, and view-specific learning objectives on LIBERO and LIBERO-Plus. The four variants of our framework use the same video sources and robot demonstrations. $G\rightarrow I$ denotes guidance from the global representation to the interaction module. Results are success rates (\%). Bold indicates the best result in each column.}
\label{tab:ablations}
\small
\setlength{\tabcolsep}{3pt}
\renewcommand{\arraystretch}{1.1}

\resizebox{\linewidth}{!}{%
\begin{tabular}{l*{5}{c}*{8}{c}}
\toprule
\multirow{2}{*}{Method}
& \multicolumn{5}{c}{LIBERO}
& \multicolumn{8}{c}{LIBERO-Plus} \\
\cmidrule(lr){2-6}
\cmidrule(lr){7-14}
& Spatial & Object & Goal & Long & Avg.
& Camera & Robot & Language & Light
& Background & Noise & Layout & Average \\
\midrule

$\pi_{0.5}$~\cite{intelligence2025pi_}
& 98.8 & 98.2 & 98.0 & 92.4 & 96.9
& 75.8 & \textbf{79.4} & 83.3 & 95.5 & 95.0 & 89.6 & \textbf{87.0} & 85.7 \\

Fast-WAM~\cite{yuan2026fast} 
& 98.2 & \textbf{100.0} & 97.0 & 95.2 & 97.6
& 17.7 & 45.3 & 60.1 & 52.2 & 22.8 & 28.5 & 61.2 & 41.3\\

Direct flat fusion + Head prediction
& 98.4 & 98.2 & 97.0 & 94.0 & 96.9
& 66.7 & 63.2 & 77.9 & 81.1 & 80.2 & 71.4 & 80.6 & 73.8 \\

Hierarchical fusion + w/o $G\to I$ guide
& 98.6 & 99.6 & \textbf{98.4} & \textbf{95.4} & 98.0
& 67.2 & 79.1 & \textbf{87.3} & \textbf{98.2} & 97.0 & 88.1 & 86.6 & 85.1 \\

Hierarchy + Head/Wrist prediction
& \textbf{99.4} & 99.6 & 98.0 & 94.4 & 97.9
& 88.8 & 70.8 & 84.9 & 96.6 & 97.4 & 92.2 & 86.1 & 87.4 \\

Hierarchy + Head prediction (Ego-CEVR)
& 99.2 & \textbf{100.0} & 98.2 & 94.8 & \textbf{98.1}
& \textbf{91.1} & 78.3 & 86.6 & 97.7 & \textbf{98.8} & \textbf{95.4} & 85.7 & \textbf{90.2} \\

\bottomrule
\end{tabular}%
}
\end{table*}

We conduct ablations on LIBERO and LIBERO-Plus to examine the hierarchical representation structure, global-to-interaction guidance, and view-specific learning objectives. The results are shown in \cref{tab:ablations}.

\paragraph{Hierarchical representation structure.}
We compare the hierarchical design with flat fusion, which combines global and wrist features through a single attention branch without the interaction module. Compared with flat fusion, \method improves the average success rate from 96.9\% to 98.1\% on LIBERO and from 73.8\% to 90.2\% on LIBERO-Plus. Under camera-viewpoint perturbations, success increases from 66.7\% to 91.1\%, a gain of 24.4 percentage points. In our hierarchical design, the global representation provides task context, while the interaction representation captures local execution cues under its guidance. The larger improvement on LIBERO-Plus suggests that this design is particularly beneficial under perturbations.

\paragraph{Global-to-interaction guidance.}
We remove the global representation from the interaction module while retaining both representations as inputs to the action expert. Without this guidance, the average success rate changes from 98.1\% to 98.0\% on LIBERO but drops from 90.2\% to 85.1\% on LIBERO-Plus. Under camera-viewpoint perturbations, success falls from 91.1\% to 67.2\%, a decrease of 23.9 percentage points. These results suggest that global-to-interaction guidance supports overall robustness to perturbations, particularly changes in camera viewpoint.

\paragraph{View-specific learning objectives.}
We extend the video prediction objective from the head view to both head and wrist views. Joint prediction achieves an average success rate of 97.9\% on LIBERO, comparable to 98.1\% for Ego-CEVR, but lowers the LIBERO-Plus average from 90.2\% to 87.4\%. One possible explanation is that wrist-camera motion and rapid changes in the visible region encourage overfitting to appearance patterns in the training data, weakening generalization.
These results support using video prediction to learn global context from head observations and robot action supervision to learn interaction representations from wrist observations.

\subsection{Knowledge Transfer on Real Robots}

\begin{wraptable}{l}{0.5\textwidth}
\centering
\caption{Real-robot comparisons and video-data ablations.
Insertion, Pouring, and Sorting refer to Plug Insertion,
Water Pouring, and Package Sorting, respectively.
All settings use the same robot demonstrations.
For Ego-CEVR, \emph{Demonstrations only} uses no additional
video data, while \emph{Extra Head} and \emph{Extra Human Ego}
add 10 h of robot head-camera and human egocentric videos,
respectively. Results are success rates (\%).}
\label{tab:ego_learning}
\small

\begin{tabularx}{\linewidth}{Lccc}
\toprule
Method & Insertion & Pouring & Sorting \\
\midrule
$\pi_{0.5}$~\cite{intelligence2025pi_} & 10 & 55 & 65 \\
Fast-WAM~\cite{yuan2026fast} & 5 & 30 & 25 \\
Demonstrations only & 20 & 60 & 60 \\
Extra Head & 35 & 95 & 85 \\
Extra Human Ego & 30 & 75 & 80 \\
\bottomrule
\end{tabularx}
\end{wraptable}

\paragraph{Learning from different video sources.}
We evaluate learning from different video sources on Plug Insertion, Water Pouring, and Package Sorting. Representative real-robot tasks and transfer settings are shown in \cref{fig:real}. We include $\pi_{0.5}$~\cite{intelligence2025pi_} and Fast-WAM~\cite{yuan2026fast} as representative VLA and WAM baselines, respectively. All methods use the same 50 robot demonstrations for policy training. For \method, \emph{Demonstrations only} uses no additional video data, while \emph{Extra Head} and \emph{Extra Human Ego} add 10 hours of robot head-camera and human egocentric videos, respectively.

As shown in \cref{tab:ego_learning}, \emph{Demonstrations only} achieves success rates of 20\%, 60\%, and 60\% on the three tasks, respectively. \emph{Extra Head} improves these rates to 35\%, 95\%, and 85\%. \emph{Extra Human Ego} achieves 30\%, 75\%, and 80\%, improving over \emph{Demonstrations only} by 10, 15, and 20 percentage points, respectively. Both settings with additional videos achieve higher success rates than $\pi_{0.5}$~\cite{intelligence2025pi_} and Fast-WAM~\cite{yuan2026fast} on all three tasks. The largest gain from human videos is observed on Package Sorting, which uses a five-finger dexterous hand, and its success rate is within five percentage points of \emph{Extra Head}. These results support transferring manipulation knowledge from human egocentric videos to robot control without additional robot demonstrations.

\paragraph{Video-guided task knowledge transfer.}
We evaluate cross-task knowledge transfer from \emph{Water Pouring} (Task 1) to \emph{Block Stacking} (Task 2) on G2. The initial training setting uses 50 G2 demonstrations of Task 1. We compare adding human egocentric videos or G2 head-camera videos of Task 2. For a fully supervised reference, we add 50 G2 demonstrations of Task 2 with action annotations.

\begin{wraptable}{l}{0.48\textwidth}
\centering
\caption{Video-guided cross-task knowledge transfer on G2.
The initial training setting uses 50 demonstrations of Water Pouring (Task 1). Additional training data for Block Stacking (Task 2) include human egocentric videos, G2 head-camera videos, or 50 action-labeled G2 demonstrations. Results are success rates (\%).}
\label{tab:task_transfer}
\small
\setlength{\tabcolsep}{3pt}

\begin{tabular*}{\linewidth}
{@{\extracolsep{\fill}}lcc@{}}
\toprule
Additional data
& G2 Task 1
& G2 Task 2 \\
\midrule
None           & 60 & 0  \\
Task 2 Human Ego & 65 & 10 \\
Task 2 G2 Head & 65 & 20 \\
Task 2 G2 Full & 75 & 55 \\
\bottomrule
\end{tabular*}
\end{wraptable}

As shown in \cref{tab:task_transfer}, the success rate on Task 2 is 0\% without additional data. Adding human egocentric videos increases this to 10\%, while G2 head-camera videos increase it to 20\%. Both video settings also raise Task 1 success from 60\% to 65\%. The fully supervised reference reaches 55\% on Task 2 and 75\% on Task 1. These results suggest that \method supports effective cross-task knowledge transfer by combining task-level information from these videos with previously learned local execution knowledge.

\begin{wraptable}{l}{0.48\textwidth}
\centering
\caption{Few-shot cross-embodiment knowledge transfer on Water Pouring (Task 1). The initial training setting uses 50 G2 demonstrations and 5 G1 demonstrations. The video settings add human egocentric videos or G1 head-camera videos. \emph{Task 1 G1 Full} retains the 50 G2 demonstrations and increases the total number of G1 demonstrations to 50. Results are success rates (\%).}
\label{tab:embodiment_transfer}
\small
\setlength{\tabcolsep}{3pt}

\begin{tabular*}{\linewidth}
{@{\extracolsep{\fill}}lcc@{}}
\toprule
Additional data
& G2 Task 1
& G1 Task 1 \\
\midrule
None           & 60 & 5  \\
Task 1 Human Ego     & 65 & 15 \\
Task 1 G1 Head & 65 & 35 \\
Task 1 G1 Full & 65 & 60 \\
\bottomrule
\end{tabular*}
\end{wraptable}

\paragraph{Few-shot cross-embodiment transfer.}
We evaluate few-shot cross-embodiment knowledge transfer from G2 to G1 on Water Pouring (Task 1). The few-shot settings use 50 G2 demonstrations and 5 G1 demonstrations of Task 1. We compare training without additional videos, with human egocentric videos, or with G1 head-camera videos of the same task. For the fully supervised reference, we retain the 50 G2 demonstrations and increase the total number of G1 demonstrations from 5 to 50.

As shown in \cref{tab:embodiment_transfer}, the success rate on G1 is 5\% without additional videos. Adding human egocentric videos increases this to 15\%, while G1 head-camera videos increase it to 35\%. Both video settings also raise G2 success from 60\% to 65\%. The fully supervised reference achieves 60\% on G1 and 65\% on G2. The 5 G1 demonstrations provide action supervision for aligning the hierarchical representations with fine-grained actions on the target robot. With this supervision, both human egocentric and G1 head-camera videos support effective few-shot cross-embodiment knowledge transfer.

\section{Discussion and Limitations}
\label{sec:limitations}

\paragraph{The need for data-native models.}
Our results suggest that egocentric videos and robot demonstrations can play complementary roles in robot learning. More broadly, how heterogeneous data are organized and learning objectives are assigned may influence how effectively models benefit from increasing data scale. As increasingly diverse datasets become available in robotics research, making effective use of their observations and supervision remains an important research challenge. These findings motivate data-native models whose learning objectives reflect the information and supervision available from each source.

\paragraph{Scaling and hierarchical training.}
As embodied data scale up, larger models are expected to learn from increasingly diverse experience. Yet how to scale learning effectively across these data sources remains an open question. This raises a further question of whether action-aligned supervision alone can lead to emergent capabilities analogous to those observed in large language models~\cite{wei2022emergent}. Even if video generation scales successfully, transferring knowledge from egocentric videos to fine-grained robot actions may still require a dedicated stage of representation learning for action alignment. We therefore believe that multistage hierarchical training will become an important direction for embodied learning.

\paragraph{Limitations and future work.}
Our framework learns from human egocentric videos and robot data, but the potential of UMI data~\cite{chi2024universal} for aligning hierarchical representations with robot actions remains unexplored. Future work will investigate whether UMI data alone can provide the required supervision, potentially reducing reliance on robot demonstrations and the associated data collection costs.
\section{Conclusion}

We present Ego-CEVR, a data-native framework that learns complementary global and interaction representations from video prediction and robot action supervision. Across simulation and real-robot experiments, Ego-CEVR achieves competitive manipulation performance and improved robustness under visual perturbations. Ablations show the benefit of hierarchical representations and of assigning different learning objectives to head and wrist observations. On real robots, additional human egocentric videos improve manipulation performance and support transfer to a new task, while few-shot adaptation enables video-learned knowledge to be aligned with fine-grained actions on a new robot. Learning from human egocentric videos and videos from other robots shares a common principle, because human-to-robot learning is itself a form of cross-embodiment knowledge transfer.

\clearpage

\bibliographystyle{unsrt}
\bibliography{ref}
\ifdefined\withsupplementary
\clearpage
\appendix
\maketitlesupplementary
\fi

\end{document}